\documentclass[runningheads]{llncs}

\usepackage{pbox}
\usepackage[11pt]{moresize}
\usepackage{changepage,threeparttable} % for wide tables
\usepackage{graphicx}
\usepackage{booktabs, multirow} % for borders and merged ranges
\usepackage{soul}% for underlines
\usepackage[table]{xcolor} % for cell colors
\usepackage{amsmath}
\usepackage{amssymb}
\usepackage{makecell}
\usepackage{hyperref}

\newcommand*\samethanks[1][\value{footnote}]{\footnotemark[#1]}

\begin{document}
\title{Improving Joint Learning of Chest X-Ray and Radiology Report by Word Region Alignment}
% \title{Improving Chest X-Ray and Radiology Report Joint Learning Using Word Region Alignment}
%
\titlerunning{Joint Learning of Chest X-Ray and Radiology Report}
\author{
Zhanghexuan Ji\inst{1}\thanks{Z. Ji and M.A. Shaikh---Equal contributions.} \and
Mohammad Abuzar Shaikh\inst{1}\samethanks{} \and
Dana Moukheiber\inst{1} \and \\
Sargur Srihari\inst{1} \and
Yifan Peng\inst{2} \and
Mingchen Gao\inst{1}
}
% index{Ji, Zhanghexuan}, index{Shaikh, Mohammad Abuzar}, index{Moukheiber, Dana}, index{Srihari, Sargur}, index{Peng, Yifan}, index{Gao, Mingchen}
%\thanks{Z. Ji and M.A. Shaikh---Equal contributions.}
\authorrunning{Z. Ji, M.A. Shaikh et al.}

\institute{
Department of Computer Science and Engineering,
University at Buffalo,\\
The State University of New York, Buffalo, NY, USA\\
\email{\{zhanghex,mshaikh2,danamouk,srihari,mgao8\}@buffalo.edu}\and
Population Health Sciences, Weill Cornell Medicine, New York, NY, USA\\
\email{yip4002@med.cornell.edu}
}
\maketitle              % typeset the header of the 
% \renewcommand{\thefootnote}{}
% \footnotetext{Z. Ji and M.A. Shaikh---Equal contributions.}
%
\begin{abstract}
Self-supervised learning provides an opportunity to explore unlabeled chest X-rays and their associated free-text reports accumulated in clinical routine without manual supervision. This paper proposes a Joint Image Text Representation Learning Network (JoImTeRNet) for pre-training on chest X-ray images and their radiology reports. The model was pre-trained on both the global image-sentence level and the local image region-word level for visual-textual matching. Both are bidirectionally constrained on Cross-Entropy based and ranking-based Triplet Matching Losses. The region-word matching is calculated using the attention mechanism without direct supervision about their mapping. The pre-trained multi-modal representation learning paves the way for downstream tasks concerning image and/or text encoding. We demonstrate the representation learning quality by cross-modality retrievals and multi-label classifications on two datasets: OpenI-IU and MIMIC-CXR. Our code is available at \url{https://github.com/mshaikh2/JoImTeR_MLMI_2021}.

%The fine-tuned encoder initialized with pre-trained representation achieves better classification performance within a shorter training time compared to supervised multi-label classification. 

\end{abstract}

\section{Introduction}

% chest x-ray background

Chest X-ray is the most common medical imaging study globally for conducting clinical routines to assess chest regions. Because of its popularity, large, labeled datasets such as ChestX-ray14 dataset~\cite{wang2017chestx}, CheXpert~\cite{irvin2019chexpert}, OpenI-IU~\cite{iu_dataset}, and MIMIC-CXR~\cite{mimicdataset,johnson2019mimic}, were collected as benchmarks for data-driven deep learning models to archive expert-level performance in analyzing chest regions. Among these biomedical datasets, OpenI-IU and MIMIC-CXR contain radiology reports along with corresponding radiographs. Given the large size of collected images and manual labeling being impractical, the disease labels are usually derived using natural language processing tools applied to the corresponding radiology reports. 

%To name a few,  ChestX-ray14 dataset \cite{wang2017chestx} collected at NIH with 14 different observation labels; CheXpert~\cite{irvin2019chexpert} collected by Stanford University with the same set of labels; OpenI-IU~\cite{iu_dataset} dataset collected from Indiana University with manual labels; and MIMIC-CXR~\cite{mimicdataset,johnson2019mimic} is the largest Chest X-ray dataset so far. Many datasets deploy automatic natural language processing tools to derive image labels in the radiology reports. Among these biomedical datasets, OpenI-IU~\cite{iu_dataset} and MIMIC-CXR~\cite{mimicdataset,johnson2019mimic} have radiology reports publicly shared with radiographs.

% self-supervised learning
Recently, self-supervised representation learning has been explored to extract underlying information from the data by performing proxy tasks that explore the organization of the data itself. This is a promising direction for learning from a large amount of unlabeled biomedical data, where manual labeling is tedious, time-consuming, subjective, and requires domain knowledge. Self-supervised learning provides a great potential to investigate the biomedical data, including both medical images and their associated reports, accumulated during clinical routines. Ideally, both the modalities of the data encode the same medical condition and should be cross-referable. 

%self attention, transformer
Self-Attention mechanism was introduced to find the cross references within the same data modality \cite{vaswani2017attention}. This concept has contributed tremendously to the recent success of natural language processing models, such as BERT \cite{devlin2019bert}. These models are pre-trained by predicting masked tokens to learn the underlying semantic representations from unlabeled textual data. Once the representation learning models are pre-trained, they can be fine-tuned and used as a backbone for a wide range of downstream natural language processing tasks. 

Motivated by the above discussion, we propose to establish the cross references of the chest radiology images and reports to jointly learn the image-text representations. Learning cross-modal visual and textual representation is an essential task that can combine the semantic information contained within images and their descriptive reports \cite{li2019visualbert,Unicoder}. These approaches have also been explored in biomedical image analysis \cite{li2020comparison}. The proposed representation learning mechanism will provide the foundation for a wide range of biomedical vision-and-language %(V+L) 
tasks, such as clinical inter-modal and intra-modal image-text retrieval, medical visual question answering \cite{abacha2019vqa}, and automatic clinical report generation \cite{li2018hybrid}.

\textbf{Contributions:} We propose JoImTeRNet - a self-supervised pre-training network trained on multimodal inputs. Our network extracts and fuses the representations of the visual and textual modalities using both global image-sentence matching and local attention-based region-phrase matching. Phrases vary from length of one to three words. The proposed local region-phrase alignment enhances the joint representation learning by performing automatic fine-grained matching between image region-of-interests with phrases in reports. The local region-phrase matching is further enhanced using a soft-attention mechanism in the image encoder, without the need for explicit manual bounding box annotation or object detection on images.
% , which in turn still heavily relies on manual bounding box labeling. 
The quality of the learned representation is tested on the downstream classification and retrieval tasks. %JoImTeRNet is trained and later the downstream encoders are fine-tuned on MIMIC-CXR dataset only. We also cross-validate the performance of these encoders on Open-I dataset without fine-tuning on it's samples.
% Furthermore, the the frozen and fine-tuned representation give better classification performance than a direct supervised classification training from scratch. The training time is also greatly reduced when the downstream task is initialized with the pre-trained representation. 

\section{Joint Image Text Representation Learning Network}
\label{sec: joimternet}
\begin{figure}[!htp]
\begin{center}
% \fbox{\rule{0pt}{2in} \rule{.9\linewidth}{0pt}}
  \includegraphics[width=0.9\linewidth]{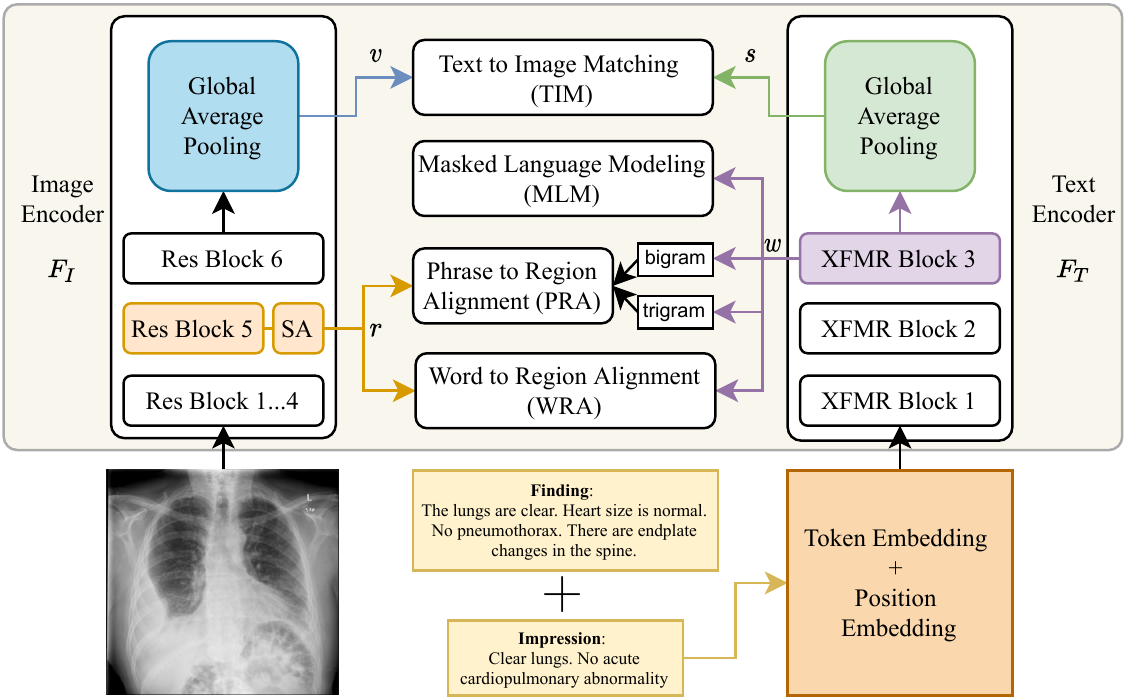}
\end{center}
   \caption{The architecture of the proposed JoImTeRNet. }
   %Feature vectors of image regions are indicated by $r$ while $v$ denotes the global image feature vector. Sentence level feature vector is indicated by $s$ while $w$ denotes the word level feature vectors.}
\label{fig:joimter_arch}
\end{figure}

We propose a \textbf{Jo}int \textbf{Im}age \textbf{Te}xt \textbf{R}epresentation Learning \textbf{Net}work (JoImTeRNet) shown in Fig. \ref{fig:joimter_arch}. The JoImTeRNet architecture consists of an image and a text encoder. The representations are matched through a list of matching tasks, Text to Image Matching (TIM), Masked Language Modeling (MLM), Phrase to Region Alignment (PRA), and Word to Region Alignment (WRA). %In addition, we introduce a phrase to region alignment task to improve the matching performance. We will discuss each of them in detail.
The learned image and text representation are mapped to a shared feature space, given the hypothesis that the radiographs and their corresponding report contain consistent semantic meaning. 

Given an X-ray image $\text{I}$ and its corresponding radiology report $\text{T}$, we first encode them with an image encoder $F_I$ and a text encoder $F_T$. The image encoder contains one input convolution layer, 6 residual blocks \cite{cnn_resnet} and a global average pooling (GAP) layer. %, where the input $\text{I}$ is down-sampled 4 times at the input layer, and is further down-sampled by 2 at each residual block. The global visual feature $v \in \mathbb{R}^{D}$ is extracted from GAP. 
In the meantime, we also get the output from the Soft-Attention (SA) \cite{shaikh2020attention} block placed after the ResBlock5 to extract the region features $r \in \mathbb{R}^{D \times M}$, such that $v,r=F_I(\text{I})$ where $v \in \mathbb{R}^{D}$ is the global image features from GAP. Sentence and word level features $s$ and $w$ are extracted using a Transformer \cite{vaswani2017attention} based text encoder $F_T$, such that $w, s = F_{T}(\text{T})$, where $w \in \mathbb{R}^{D \times N}$ and $s \in \mathbb{R}^{D}$. Three transformer layers are deployed in $F_T$ to encode the text report with the self-attention mechanism. 

% We define the $F_T$ similar to \cite{vaswani2017attention}: Given text $T\in \mathbb{R}^{N}$, we use word-based token to embed it as $e\in \mathbb{R}^{D\times N}$, which is summed with positional encodings as the input of $F_T$. In the transformer layer, $e$ is first transformed into queries $Q=W^T_Qe$, keys $K=W^T_Ke$ and values $V=W^T_Ve$ within each attention head, where $W_Q, W_K, Q_V\in \mathbb{R}^{D\times D_k}$. Then ``scaled dot-product self-attention'' is applied to $e$ as follows:
% \begin{equation}
%     \text{Attention}(Q,K,V)=\text{Softmax}(\frac{QK^T}{\sqrt{D_{k}}})V
% \label{eq:attenc_head}
% \end{equation}
% % where $f_s$ is the softmax function. 
% Multi-head attention is applied to the self-attention sub-layer and the outputs from $h$ heads are concatenated: 
% \begin{equation}
%     \text{Multihead}(e) = W_O\text{Concat}(\text{head}_1,...,\text{head}_h)
% \label{eq:attenc_multihead}
% \end{equation}
% where $W_O\in \mathbb{R}^{D\times hD_k}$ and $\text{head}_i = \text{Attention}(W^T_{Q_i}e,W^T_{K_i}e,W^T_{V_i}e)$. 

% The output from the multi-head attention is then sent to a feed-forward network. The residual mechanism \cite{cnn_resnet} and layer normalization are applied to multi-head attention and FFN outputs: 
% \begin{equation}
% \begin{aligned}
%     \Tilde{e} &= \text{LayerNorm}(e+\text{Multihead}(e)) \\
%     w &= \text{LayerNorm}(\Tilde{e}+\text{FFN}(\Tilde{e}))
% \end{aligned}
% \label{eq:attenc_output}
% \end{equation}
% and $s = \bar{w}$ is used as the sentence representation feature. 

\subsection{Matching Images and Sentences}
\label{sub: MIT}

To learn the joint representation of image and text pairs, we use the cross-entropy based matching (CEM) loss \cite{xu_attngan} and the ranking-based triplet matching (TM) loss \cite{chauhan2020joint}. Given a batch of image-text pairs ${(\text{I}_i,\text{T}_i)}_{i=1}^B$ (B is the batch size) and their corresponding visual features $v$ and sentence features $s$ from $F_I$ and $F_T$, %we calculate the posterior probability of $\text{T}_i$ matching with $\text{I}_i$ using softmax as:
the image-to-text CEM loss $L_{CEM}^{\text{IT}}$ is defined as the negative log posterior probability of the images being matched with their corresponding texts, \textit{i.e.},
% \vspace{-0.5em}
\begin{equation}
    L_{CEM}^{\text{IT}} = -\sum_{i=1}^B log(P(\text{T}_i|\text{I}_i)) 
    = -\sum_{i=1}^B log(\frac{e^{\gamma S(\text{I}_i,\text{T}_i)}}{\sum_{j=1}^Be^{\gamma S(\text{I}_i,\text{T}_j)}})%P(\text{T}_i|\text{I}_i) = \text{Softmax}_T(\gamma S(I_i,T_i)) = \frac{e^{\gamma S(\text{I}_i,\text{T}_i)}}{\sum_{j=1}^Be^{\gamma S(\text{I}_i,\text{T}_j)}}% 
%\label{eq:prob_sent}
\label{eq:loss_sent}
\end{equation}
where $\gamma$ is the smoothing factor. $P(\text{T}_i|\text{I}_i)$ is the posterior probability of $\text{T}_i$ matching with $\text{I}_i$ using softmax. Cosine similarity $S(\text{I}_i,\text{T}_i)=(v^T s)/(\Vert v\Vert\Vert s\Vert)$ is used as the similarity score between image-text pairs. During the training, $\text{T}_i$ is the correct match to $\text{I}_i$ in the batch and all the other $\text{T}_{j}(j\neq i)$ are mismatching texts. %Then, the negative log posterior probability is used to calculate CEM loss $L_{CEM}^{\text{IT}}$ as:
% \vspace{-0.5em}
% \begin{equation}
%     L_{CEM}^{\text{IT}} = -\sum_{i=1}^B log(P(\text{T}_i|\text{I}_i))
% \label{eq:loss_sent}
% \end{equation}
Considering that image-text joint representation mapping should be bidirectional, we reverse $\text{I}$ and $\text{T}$ in Eq.\eqref{eq:loss_sent} and get the symmetric text-to-image CEM loss as $L_{CEM}^{\text{TI}}$. Thus, the bidirectional CEM loss for globally matching image and text is defined as $L_{CEM}^s=L_{CEM}^{\text{IT}}+L_{CEM}^{\text{TI}}$.

Although CEM loss is designed to make the similarity between correct image-text pairs relatively higher than other mismatched pairs, it is difficult to set a hard margin between mismatched features. To solve this problem, TM loss \cite{chauhan2020joint}, a ranking-based criterion,  is added to increase the distance of mismatched pairs in the joint embedding space. Given an image $\text{I}_i$ as the anchor, $\text{T}_i$ is used as the positive paired sample. We then randomly select a mismatching text $\text{T}_{j} (j\neq i)$ within the batch as the negative paired sample. Symmetrically, if $\text{T}_i$ is used as the anchor, then $\text{I}_i$ and $\text{I}_{j}$ would be positive/negative samples. The bidirectional TM loss for global image-text matching is formed as:
\begin{equation}
\begin{aligned}
    L_{TM}^s = L_{TM}^{\text{IT}} + L_{TM}^{\text{TI}}
    & = \sum_{i=1}^{B}\Big[\text{max}(0,S(\text{I}_i,\text{T}_{j})-S(\text{I}_i,\text{T}_i)+\eta_s) \Big.\\ 
    & \Big.\quad\quad\quad + \text{max}(0,S(\text{I}_{j},\text{T}_i)-S(\text{I}_i,\text{T}_i)+\eta_s)\Big]
\end{aligned}
\label{eq:triplet_sent}
\end{equation}
where $\eta_s$ is the hard margin and $S$ is the cosine similarity the same as in Eq.\eqref{eq:loss_sent}. 

\subsection{Aligning Image Regions and Report Phrases}
\label{sub: AIRW}
Both chest X-rays and their corresponding reports contain lots of fine-grained %detailed
semantic information. %. Given such rich semantic information
%, which suggests that simply matching image and text with the global embedded feature may not be %efficient
%sufficient. Thus, 
We introduce a region-phrase level matching to align different concepts in the text reports with the regions of the images to further improve the joint representation. We apply region-phrase alignment with both CEM loss and TM loss. The length of a phrase is in the range of 1 to 3 words. Features of words, bigram and trigram phrases are denoted as $w, p_2, p_3$ respectively.

%Since the mapping between regions and words/phrases are not explicit, we cannot directly use cosine similarity to calculate matching scores %between regions and words
%in this case. 
The cosine similarity between regions and words/phrases is not feasible to calculate directly due to the lack of explicit mapping between them. Instead, an attention-based matching score is deployed to overcome this challenge \cite{fang2015captions,huang2013learning,xu_attngan}. For region-word-level matching, given $(\text{I}_i,\text{T}_i)$ and their region-word features $(r,w)$, 
we first calculate the similarity matrix between all possible pairs of region features and word features using dot-product, \textit{i.e.}, $m=w^Tr$, 
%the word-region attention score $\alpha$ is computed using dot-product.
% \vspace{-0.5em}
% \begin{equation}
%     m=w^Tr
%     \vspace{-0.5em}
%     \label{eq:sim_matrix}
% \end{equation}
%Similarity between all possible pairs of word features and region features is captured in $m$, 
where $m\in\mathbb{R}^{N\times M}$, which is further normalized %via softmax 
along $N$ words as $\bar{m}=\text{Softmax}_N(m)$. %Next $\alpha$ is multiplied with the region features $r$ to obtain the context vector $c$ as follows:
Next, a context feature $c$ is computed as the weighted sum over region features $r$, weighted by the region-word attention score $\alpha$ as follows:
% \vspace{-0.5em}
\begin{equation}
\begin{aligned}
    c = \alpha r^T \text{, where } \alpha_{i,j} = \frac{e^{\gamma_1 \bar{m}_{i,j}}}{\sum_{k=0}^{M-1} e^{\gamma_1 \bar{m}_{i,k}}} %\text{, } \bar{m} = \frac{e^{m_{i,j}}}{\sum_{k=0}^{N-1} e^{m_{k,j}}}
\end{aligned} 
\label{eq:context_aware_vector}
\end{equation} 
where $c \in \mathbb{R}^{N\times D}$ and $\alpha \in \mathbb{R}^{N\times M}$; $\gamma_1$ is a hyper-parameter to tune the required amount of visual attention for a word. Here, the $i^{th}$ vector of $c$ is the attention-weighted representation of all the sub-regions related to the $i^{th}$ word. 

The attention-based region-word-level matching score is computed as: %$S_a$ for $(\text{I},\text{T})$ as:
% \vspace{-0.5em}
\begin{equation}
    S_a(\text{I},\text{T}) = \log(\sum_{i=1}^{N-1}e^{(\gamma_{2}S(c_i,w_i))})^{\frac{1}{\gamma_2}}
\label{eq:attention_similarity}
\end{equation}
where $S(c_i,w_i)=(c_i^T w_i)/(\Vert c_i\Vert\Vert w_i\Vert)$, is the element-wise cosine similarity score between $c_i$ and $w_i$, $\gamma_2$ is the importance magnification hyper-parameter for the most relevant word and context vector pair. 

By replacing the cosine similarity score $S(\cdot,\cdot)$ with the region-word matching score $S_a(\cdot,\cdot)$ in Eq.\eqref{eq:loss_sent}\eqref{eq:triplet_sent}, we obtain the bidirectional CEM loss and TM loss for region-word alignment as $L_{CEM}^{p_1}=L_{CEM}^{rw}+L_{CEM}^{wr}$ and $L_{TM}^{p_1}=L_{TM}^{rw}+L_{TM}^{wr}$. 

Furthermore, we obtain the phrase features by applying a 1D convolutional layer with kernel size 2 and 3 over $w$ to get bigram $p_2 = \theta_{p_2}^Tw$ and trigram $p_3 = \theta_{p_3}^Tw$ phrase features respectively \cite{kim-2014-convolutional,yang2016stacked}. Here, $\theta_{p_2}, \theta_{p_3}$ are the convolution kernels of size 2 and 3.
%By substituting $p_2$ and $p_3$ for $w$ when calculating the similarity matrix $m$ in Eq.\eqref{eq:context_aware_vector} as well as the similarity score in Eq.\eqref{eq:attention_similarity}, we obtain the bidirectional bigram/trigram losses. 
%as $L_{CEM}^{p_2} = L_{CEM}^{rp_{2}}+L_{CEM}^{p_{2}r}$ and $L_{TM}^{p_2} = L_{TM}^{rp_{2}}+L_{TM}^{p_{2}r}$ and trigram losses as $L_{CEM}^{p_3} = L_{CEM}^{rp_{3}}+L_{CEM}^{p_{3}r}$ and $L_{TM}^{p_3} = L_{TM}^{rp_{3}}+L_{TM}^{p_{3}r}$.
Our final cross-entropy with triplet matching (CETM) loss for our image-text joint representation learning is designed as:
% \vspace{-0.5em}
\begin{equation}
    L_{CETM} = \lambda_{CEM}( L_{CEM}^s + \sum_{i=1}^{3}L_{CEM}^{p_{i}}) + \lambda_{TM}( L_{TM}^s + \sum_{i=1}^{3}L_{TM}^{p_{i}})
\label{eq:final_loss}
\end{equation}
where $\lambda_{CEM}$ and $\lambda_{TM}$ are the loss weight hyper-parameters.

\subsection{Downstream Task}
\label{sub: CLS}
In order to demonstrate the performance of joint representation learning, we use the pre-trained image and text encoders as the backbone and test the learned features on multi-label classification. We add projection layers followed by two fully connected layers for multi-label classification. %The features $r,v,w,s$ are projected, pooled and then concatenated as the input to the fully connected layers% to get the classification prediction. 
Cross-entropy loss balanced with positive/negative ratio and class-wise weights \cite{wang2018tienet} are used for training.

\section{Experiments}

% In this section, we explain our experiment settings, evaluation metrics and results to evaluate our proposed method. 

\subsection{Datasets} 
% Our model is evaluated on two public datasets including OpenI's Indiana University Chest X-ray Collection (OpenI-IU) \cite{iu_dataset} and MIMIC-CXR v2.0 \cite{mimicdataset}.

\textbf{MIMIC-CXR} v2.0 \cite{mimicdataset}, is a large public dataset consisting of 377,110 chest X-rays associated with 227,835 radiology reports. %13 disease labels are extracted from reports by auto-annotator. 
We limit our study to the frontal-view images and only keep one frontal view image for each report. Following the pre-processing scheme in \cite{chauhan2020joint}, we extract the \textit{impressions, findings, conclusion} and \textit{recommendation} sections from the raw report, normalized by SciSpaCy \cite{neumann2019scispacy}, and concatenate them. If none of these sections are present, we use the \textit{final report} section. 14 CheXpert labels provided in MIMIC-CXR are used for classification task, where label 1 is considered as positive and all the other labels (-1, 0) and missing labels are merged as negative. This results in 222,252 image-report pairs with 14 binary labels. We split the dataset into 217,252, 2,000 and 3,000 samples for training, validation and testing respectively.

\noindent \textbf{OpenI-IU} \cite{iu_dataset} is a public dataset with 3,996 radiology reports and 8,121 associated chest X-ray images, which are manually annotated by human experts using MeSH words. Similar to TieNet \cite{wang2018tienet}, only unique frontal images and their corresponding reports which contain either \textit{findings} and/or \textit{impressions} are selected. This yields 3,643 image-report pairs, which are only used as external evaluation sets. For comparison and evaluation purposes, we select the 7 common labels in both OpenI and MIMIC-CXR (Table \ref{tab: classification_openi}) from the MeSH domain. 
%The descriptions remain the same for patients who have more than one frontal images. 
%For each report we randomly select one frontal image during the training. 
%We further split this collection to 3,166, 200, and 300 data-points containing image-text pairs for training, validation and testing respectively. For the downstream classification, we extract the labels from the  ``Problems'' section of the report. As the dataset is very small and imbalanced, we consider only the labels which have more than 50 data-points as our ``Disease" class and all other labels are considered as ``No Finding'', which means the selected diseases are not present.

% \newline
% \textbf{Image pre-processing:} For both OpenI-IU and MIMIC-CXR the input image is cropped or padded to $2048 \times 2048$. The image is normalized so that its pixel values range between -1 and 1. Random crop, rotation and color jitter (contrast, brightness, saturation) are further used for data augmentation during training. 

% \newline
% \textbf{Text pre-processing:} For OpenI-IU dataset we concatenate the \textit{findings} and \textit{impressions} sentences for each image. Furthermore, we apply a word-level tokenization scheme, collect only the words which appear more than twice in the dataset and reduce the vocabulary size to $1,196$ and $8,410$ for OpenI-IU and MIMIC-CXR respectively. The sentences are truncated or padded to the max length of $N=160$ and $N=256$ for OpenI-IU and MIMIC-CXR respectively. 
\begin{table}[!tp]
\centering
\caption{Ablation for selecting the best loss setting. The matching score for OpenI-IU and MIMIC-CXR is computed on 1000 and 1000/3000 test samples respectively. Subscript $s,w,p$ stand for image-text, region-word and region-phrase level matching.}
\resizebox{1.0\textwidth}{!}{%
% \scriptsize
\begin{tabular}{l|ccc|ccc|ccc|ccc|ccc|cccc}\toprule
\multirow{3}{*}{\makecell{Model\\ setting}
} &\multicolumn{12}{c}{MIMIC-CXR} &\multicolumn{6}{|c}{OPENI-IU} \\\cmidrule{2-19}
&\multicolumn{3}{c}{I2T (1K)} &\multicolumn{3}{|c}{T2I (1K)} &\multicolumn{3}{|c}{I2T (3K)} &\multicolumn{3}{|c}{T2I (3K)} &\multicolumn{3}{|c}{I2T (1K)} &\multicolumn{3}{|c}{T2I (1K)} \\\cmidrule{2-19}
&R@1 &R@5 &R@10 &R@1 &R@5 &R@10 &R@1 &R@5 &R@10 &R@1 &R@5 &R@10 &R@1 &R@5 &R@10 &R@1 &R@5 &R@10 \\\midrule
% TML\textsubscript{s}
TM\textsubscript{s}\cite{chauhan2020joint} &5.37 &19.43 &30.73 &5.40 &20.23 &30.23 &2.37 &9.70 &15.63 &2.23 &10.20 &16.37 &1.83 &5.70 &9.13 &1.50 &5.67 &9.30 \\
TM\textsubscript{ws} &6.30 &21.73 &32.23 &6.00 &20.97 &30.90 &2.77 &10.67 &18.07 &2.83 &10.97 &17.83 &1.93 &6.37 &10.17 &1.97 &6.53 &10.30 \\
% CEML\textsubscript{s}huang2013learning,
CEM\textsubscript{s}\cite{fang2015captions} &18.60 &43.10 &56.10 &18.13 &43.20 &55.97 &12.20 &31.27 &41.80 &11.80 &30.87 &41.10 &4.70 &12.83 &17.73 &4.83 &12.30 &17.87 \\
% CEML\textsubscript{ws}
CEM\textsubscript{ws}\cite{xu_attngan} &18.60 &44.20 &56.27 &18.87 &43.40 &55.67 &12.60 &31.67 &41.80 &12.83 &31.57 &41.27 &4.87 &13.00 &18.00 &5.37 &13.40 &18.33 \\
CETM\textsubscript{ws} &\textbf{19.07} &45.33 &57.20 &\textbf{19.07} &44.70 &56.73 &\textbf{12.77} &31.90 &43.00 &\textbf{12.97} &31.97 &42.03 &\textbf{5.13} &\textbf{13.07} &18.73 &5.50 &13.73 &19.20 \\
CETM\textsubscript{wps} &18.93 &\textbf{46.20} &\textbf{58.67} &\textbf{19.07} &\textbf{45.27} &\textbf{58.50} &12.67 &\textbf{33.20} &\textbf{44.07} &12.83 &\textbf{32.43} &\textbf{43.40} &5.07 &\textbf{13.07} &\textbf{18.83} &\textbf{5.67} &\textbf{13.83} &\textbf{19.20} \\
\bottomrule
\end{tabular}}
\label{tab: ablation}
% \vspace{-3em}
\end{table}
\subsection{Implementation Details}

\label{sub: implementation}

JoImTeRNet is implemented in Pytorch \cite{torch2019} and all the experiments are carried out on NVIDIA GTX 1080 Ti GPUs. For $F_I$, we use the basic residual blocks proposed in \cite{cnn_resnet}. We employ 3 layers of Transformer blocks with 8 heads in $F_T$. The input image is encoded into 256 regions ($r$) flattened from $16 \times 16$ feature map output from Res Block 5 as shown in Fig. \ref{fig:joimter_arch}. The input image is cropped or padded to $2048 \times 2048$ then normalized to [-1, 1]. Random crop, rotation and color jitter are used for data augmentation. Report input is tokenized by a word-level tokenization scheme, where we collect all the words that appear more than twice in MIMIC-CXR dataset, which results in vocabulary size of $8,410$. The input reports are truncated or padded to the max length of $N=160$. 

\noindent \textbf{Parameter Settings} We pretrain $F_I$ and $F_T$ on MIMIC-CXR training set using the image-text matching task explained in Section \ref{sub: MIT} and \ref{sub: AIRW} to generate the joint image and text representations. The maximum epoch is set as $30$. We employ AdamW \cite{adam2014} optimizer with an initial learning rate of $10^{-4}$, which is dropped by $10$ times after $20$ epochs. L2 weight decay is set as $10^{-4}$. %We monitor the CEM and TM Loss curves and stop the training if the validation losses overfit or plateau. 
For the downstream classification task in Section \ref{sub: CLS}, we set up two different settings for comparison: randomly initializing the backbone and fine-tuning the pre-trained backbone. The learning rate for the classification head in both settings is set to $10^{-4}$. For the randomly initialized setting,  we train the backbone using the same learning rate as the classification block for 20 epochs, whereas the pre-trained backbone is fine-tuned with a smaller learning rate of $10^{-5}$ for only 10 epochs. Our model pre-trained with the full loss setting CETM\textsubscript{wps} is used as the backbone for fine-tuning. 
The batch size is set to 32 for all the experiments. We select the loss hyper-parameters as $\gamma,\gamma_1,\gamma_2=2,1,1$, $\eta_s, \eta_w=0.5,0.5$ and $\lambda_{CEM},\lambda_{TM}=2.0, 1.0$.
%(detailed  in supplementary Table \ref{supptab: hyperparameter})

\begin{table}[!tp]\centering
\caption{Classification AUCs on MIMIC-CXR \cite{mimicdataset} dataset. ``FS" stands for training from scratch (FS). ``FT" stands for fine-tuned model. Other comparison experiments are Visualbert \cite{li2019visualbert}, Uniter \cite{chen2020uniter}, and ClinicalBert \cite{alsentzer2019publicly}.
}
% \caption{Generated by Spread-LaTeX}\label{tab: }
% \scriptsize
\resizebox{0.7\textwidth}{!}{%
\begin{tabular}{l|cc|cccc|cccc}\toprule
\multirow{2}{*}{Findings} &\multicolumn{2}{c}{Image} &\multicolumn{4}{|c}{Image+Report} &\multicolumn{3}{|c}{Report} \\\cmidrule{2-10}
&FS &FT &\cite{li2019visualbert} &\cite{chen2020uniter} &FS &FT &\cite{alsentzer2019publicly} &FS &FT \\\midrule
EC &0.738 &\textbf{0.763} &0.981 &0.979 &0.989 &\textbf{0.996} &0.966 &0.986 &\textbf{0.980} \\
Cardiomegaly &0.794 &\textbf{0.820} &0.991 &0.989 &0.989 &\textbf{0.993} &0.979 &0.988 &\textbf{0.989} \\
Airspace Opacity &0.749 &\textbf{0.759} &0.991 &0.989 &0.988 &\textbf{0.992} &0.978 &0.985 &\textbf{0.986} \\
Lung Lesion &0.696 &\textbf{0.756} &0.985 &0.981 &0.996 &\textbf{0.998} &0.972 &\textbf{0.989} &0.987 \\
Edema &0.883 &\textbf{0.895} &0.991 &0.990 &0.995 &\textbf{0.996} &0.979 &0.993 &\textbf{0.994} \\
Consolidation &0.794 &\textbf{0.810} &0.989 &0.988 &0.994 &\textbf{0.998} &0.979 &\textbf{0.996} &0.995 \\
Pneumonia &0.733 &\textbf{0.738} &0.977 &0.974 &0.981 &\textbf{0.985} &0.962 &0.980 &\textbf{0.984} \\
Atelectasis &0.808 &\textbf{0.824} &0.988 &0.987 &0.988 &\textbf{0.995} &0.976 &0.991 &\textbf{0.992} \\
Pneumothorax &0.845 &\textbf{0.855} &0.992 &0.991 &0.989 &\textbf{0.993} &0.979 &0.993 &\textbf{0.994} \\
Pleural Effusion &0.898 &\textbf{0.904} &0.993 &0.992 &0.986 &\textbf{0.997} &0.981 &0.989 &\textbf{0.990} \\
Pleural Others &0.812 &\textbf{0.839} &0.981 &0.973 &0.996 &\textbf{0.999} &0.964 &\textbf{0.998} &0.993 \\
Fracture &0.641 &\textbf{0.714} &0.976 &0.977 &\textbf{0.997} &0.990 &0.958 &\textbf{0.997} &\textbf{0.997} \\
Support Devices &0.901 &\textbf{0.913} &\textbf{0.995} &0.994 &0.992 &\textbf{0.995} &0.983 &0.988 &\textbf{0.993} \\
No Findings &0.865 &\textbf{0.874} &- &- &0.985 &\textbf{0.989} &- &\textbf{0.980} &\textbf{0.980} \\
\cline{1-10}
Avg &0.792 &\textbf{0.815} &0.987 &0.985 &0.991 &\textbf{0.994} &0.974 &\textbf{0.990} &\textbf{0.990} \\
\bottomrule
\end{tabular}
}
\label{tab: classification_mimic}
\end{table}
% \vspace{-0.1in}

\subsection{Performances}
% \vspace{-0.08in}
\textbf{Evaluation Metric} We evaluate the performance of JoImTeRNet by cross-modality retrieval task: given one image (text) as a query, we rank a subset of text (image), including the paired one, based on cosine similarity between the image and text features from JoImTeRNet. Recall@K (R@K) \cite{karpathy2015deep} is reported, where $K \in \{1,5,10\}$, which measures the fraction of times the correct matching is retrieved among the top $K$ results in the test set. We compute R@K on a subset of 1000 image-text pairs and on the full 3000 samples in our MIMIC-CXR test set. We also report R@K on a subset of 1000 samples in OpenI-IU in order to evaluate JoImTeRNet on the external dataset (Table \ref{tab: ablation}). %For the downstream multi-label classification, we report the area under ROC (AUC) in Table \ref{tab: classification_mimic} and Table \ref{tab: classification_openi}. For OpenI-IU, as the test set contains only 300 samples, we compute R@K on a batch of 100 samples as in Table \ref{tab: ablation}. 

\noindent\textbf{Ablation Study for Loss Settings} Ablation studies for different combinations of our losses are listed in Table \ref{tab: ablation}. As we can see, full loss setting CETM\textsubscript{wps} achieves the highest R@5 and R@10 scores on all the test set, %and improves the scores by about 1\% compared with CETM\textsubscript{ws}, 
which shows the effectiveness of our multilevel phrase matching loss.  In addition, the matching performance degrades when the model is trained on global matching loss only without the region-phrase(word)-level matching, i.e. CEM\textsubscript{s} performs worse than CEM\textsubscript{ws}. Similar results are found when comparing TM\textsubscript{s} with TM\textsubscript{ws}. This result shows that our proposed method for assisting joint representation learning using region-word matching is able to improve the representation ability of the image-text encoder. Moreover, the CETM combination consistently gains performance compared with only CEM loss or TM loss settings, which is just as we expected in Sections \ref{sub: MIT} and \ref{sub: AIRW}. Notice that the matching scores are much lower on OpenI, since OpenI contains a large amount of similar reports, e.g. `No acute disease.', which can have very similar feature representation from our model and thus largely degrade the matching score. %The decrease in performance when no region-word-level matching loss is used proves that      Also, with a larger dataset, the performance difference is reduced, probably due to the model's ability to learn more details from global matching itself when a large and diverse training set is used. Another interesting finding is that, CEM gives a much better performance than TM on MIMIC. This can be explained as TM loss only focuses on the distance between positive and negative pairs within one triplet each time, therefore it loses the full insight on all the negative samples within each batch. At the same time, CEM uses softmax over all possible pairs, which provides greater information to supplement model learning. 

% Some R@1 scores of CETM\textsubscript{wps} are slightly lower then CETM\textsubscript{ws}, which might due to the phrase convolution .

\begin{table}[!bt]\centering
\caption{
Classification AUCs on OpenI-IU \cite{iu_dataset} dataset. ``FS" stands for training From Scratch (FS). ``FT" stands for Fine-tuned model. Other comparison experiments are ChestX-ray14 \cite{wang2017chestx}, TieNet  \cite{wang2018tienet}, Visualbert \cite{li2019visualbert}, Uniter \cite{chen2020uniter}, and ClinicalBert \cite{alsentzer2019publicly}.
}
% \caption{Generated by Spread-LaTeX}\label{tab: }
% \scriptsize
\resizebox{\textwidth}{!}{%
\begin{tabular}{l|ccc|ccccc|cccc|cc}\toprule
\multirow{2}{*}{Findings} &\multicolumn{3}{c}{Image} &\multicolumn{5}{|c}{Image+Report} &\multicolumn{4}{|c|}{Report} &\multirow{2}{*}{\makecell{No. of \\ Samples}} \\\cmidrule{2-13}
&\cite{wang2017chestx} &FS &FT & \cite{wang2018tienet} & \cite{li2019visualbert} &\cite{chen2020uniter} &FS &FT &\cite{wang2018tienet} &\cite{alsentzer2019publicly} &FS &FT & \\\midrule
Cardiomegaly &0.803 &0.924 &\textbf{0.937} &0.962 &0.977 &0.978 &0.956 &\textbf{0.985} &0.944 &0.969 &0.966 &\textbf{0.987} &315 \\
Edema &0.799 &0.937 &\textbf{0.953} &\textbf{0.995} &0.982 &0.989 &0.922 &0.962 &\textbf{0.984} &0.976 &0.947 &0.964 &40 \\
Consolidation &0.790 &0.951 &\textbf{0.951} &0.989 &0.996 &\textbf{0.998} &0.954 &0.975 &0.969 &\textbf{0.982} &0.938 &0.975 &28 \\
Pneumonia &0.642 &0.863 &\textbf{0.934} &\textbf{0.994} &0.990 &0.988 &0.877 &0.949 &\textbf{0.983} &0.982 &0.880 &0.943 &36 \\
Atelectasis &0.702 &0.829 &\textbf{0.858} &0.972 &\textbf{0.992} &0.982 &0.947 &0.978 &\textbf{0.981} &0.947 &0.952 &0.971 &293 \\
Pneumothorax &0.631 &0.926 &\textbf{0.936} &0.960 &0.988 &0.983 &0.962 &\textbf{0.989} &0.960 &0.973 &0.951 &\textbf{0.989} &22 \\
Pleural Effusion &0.890 &0.938 &\textbf{0.957} &0.977 &\textbf{0.985} &0.983 &0.922 &0.971 &0.968 &\textbf{0.976} &0.926 &0.968 &140 \\
No Finding &- &0.844 &\textbf{0.851} &- &- &- &0.883 &\textbf{0.961} &- &- &0.898 &\textbf{0.930} &2789 \\
\cline{1-14}
Avg &0.751 &0.910 &\textbf{0.932} &0.978 &\textbf{0.987} &0.986 &0.934 &0.973 &0.970 &\textbf{0.972} &0.932 &0.971 & \\
\cline{1-14}
W. Avg &0.771 &0.893 &\textbf{0.915} &0.971 &\textbf{0.985} &0.982 &0.943 &0.978 &0.965 &0.964 &0.949 &\textbf{0.975} & \\
\bottomrule
\end{tabular}
}
\label{tab: classification_openi}
\end{table}
%\noindent \textbf{Negation Pattern Detection} Negation is a major source of errors since many of the findings describe their absence rather than presence. Although our method is not designed to handle the negation explicitly, we evaluate our learned text features on detecting negation patterns on the manually labeled reports on MIMIC-CXR dataset. The results show that our learned features are able to detect negation patterns compared to the NegBio \cite{peng2018negbio} (Detailed in supplementary Table S2).

\noindent\textbf{Downstream Image Classification Results} 
The AUCs from our two settings on both datasets along with other SOTA performances are shown in Table \ref{tab: classification_mimic} and \ref{tab: classification_openi}. We can see that the classifier performance finetuned on JoimTerNet backbone (FT) is always higher than training from scratch (FS), which shows the advance of our pre-training method. As shown in Table \ref{tab: classification_mimic}, FT achieves the highest AUCs on most tasks and labels on MIMIC-CXR test set (internal evaluation), even better than some SOTA models \cite{chen2020uniter,li2019visualbert,alsentzer2019publicly} on image-text and text classification. 
For the external evaluation on OpenI in Table \ref{tab: classification_openi}, our FT setting extremely improves average AUC on image classification by 18\% compared with TieNet \cite{wang2018tienet}, and also gains 1\% on wAvg AUC than ClinicalBERT \cite{alsentzer2019publicly} on report classification. For the image-text classification, our model is still comparable with other SOTA models, even though our text encoder only contains 3 transformer layers compared with \cite{chen2020uniter,li2019visualbert} which has a 12 layer BERT encoder as the backbone. 
% The AUC results from our three training strategies on both datasets are shown in Table. \ref{tab: classification}. As we can see, both settings initialized with our pretrained backbone generate better classification results on wAvg-AUC and the AUCs for all the labels, especially for the large MIMIC dataset, where most of the AUCs are improved by 1-3\% and the wAvg-AUC is boosted above 84\% compared with random initialization from scratch. Some hard labels like `Fracture' and `Lung Lesion' are largely increased by 6\%. This result suggests that our self-supervised image-text representation pretraining successfully extracts the important medical information within its image embedded features for the downstream task, even without finetuning. Additionally, finetuning the pretrained backbone results in slightly higher classification accuracy than the frozen setting on both datasets, which means that we should always finetune the pretrained model with a small learning rate in the downstream tasks for a better performance. Moreover, using the pretrained backbone significantly reduces the training time by 60\%. On the original MIMIC data, it takes more than 25 epochs for the model to converge, whereas with our pretrained image encoder, the convergence time drops to 10 epochs. 

\section{Conclusion}
We propose a joint image-text representation learning network and show its performance on cross-modality retrieval and multi-label classification. We demonstrate the potential of self-supervised learning when it meets the continuously generated biomedical images and reports. We also leverage and show the importance of information contained within the relationship of words, phrases and image regions. Future work includes more complicated downstream tasks regarding both images and text.
%, for instance, medical VQA and automatic report generation.

\textbf{Acknowledgments} 
  This research was supported in part by NSF through grant IIS-1910492.  It also was supported by the National Library of Medicine under Award No. 4R00LM013001.

\bibliographystyle{splncs04}
% \bibliography{References_MMRL, References_Dana, References_Mingchen}
\bibliography{paper20}

\end{document}